\documentclass[letterpaper, 10 pt, conference]{ieeeconf}  

\IEEEoverridecommandlockouts                              

\usepackage{graphics} 
\usepackage{epsfig} 
\usepackage{amssymb}  
\usepackage{bm}
\usepackage{algpseudocode}
\usepackage{algorithm}
\usepackage{xcolor}
\usepackage{amsmath}
\usepackage{hyperref}
\usepackage{multirow}
\usepackage{booktabs}

\usepackage{caption}
\usepackage{subcaption}

\usepackage{bbm}

\title{\LARGE \bf
ViPFormer: Efficient Vision-and-Pointcloud Transformer for Unsupervised Pointcloud Understanding
}

\author{Hongyu Sun, Yongcai Wang, Xudong Cai, Xuewei Bai and Deying Li
  \thanks{All authors are with the Department of Computer Science, School of Information, Renmin University of China, 
                Beijing 100872, China. Corresponding author: Yongcai Wang. 
                {\tt\small \{sunhongyu, ycw, xudongcai, bai\_xuewei, deyingli\}@ruc.edu.cn} }%
  \thanks{This work was supported in part by the National Natural Science Foundation of China under Grants 
  No. 61972404 and No. 12071478, and Public Computing Cloud, Renmin University of China.}%
}

\begin{document}

\maketitle
\thispagestyle{empty}
\pagestyle{empty}

\begin{abstract}

Recently, a growing number of work design unsupervised paradigms for point cloud processing to alleviate the limitation of 
expensive manual annotation and poor transferability of supervised methods. 
Among them, CrossPoint follows the contrastive learning framework and exploits 
image and point cloud data for unsupervised point cloud understanding. Although the promising performance is presented, 
the unbalanced architecture makes it unnecessarily complex and inefficient. 
For example, the image branch in CrossPoint is $\sim$8.3x heavier than the point cloud branch leading to higher 
complexity and latency. To address this problem, in this paper, we propose a lightweight Vision-and-Pointcloud Transformer (ViPFormer) 
to unify image and point cloud processing in a single architecture. ViPFormer learns in an unsupervised manner by optimizing 
intra-modal and cross-modal contrastive objectives. Then the pretrained model is transferred to 
various downstream tasks, including 3D shape classification and semantic segmentation. 
Experiments on different datasets show ViPFormer surpasses previous state-of-the-art unsupervised methods with higher accuracy, 
lower model complexity and runtime latency. Finally, the effectiveness of each component in ViPFormer is validated by
extensive ablation studies. The implementation of the proposed method is available at 
\url{https://github.com/auniquesun/ViPFormer}.

\end{abstract}

\section{Introduction}
\label{sec:intro}

Point cloud understanding is a crucial problem which
has attracted widespread attention for its values in autonomous driving, mixed reality, and robotics.
There are three common tasks in point cloud understanding: 3D object classification~\cite{xiang21curvenet,ma22pointmlp}, 
semantic segmentation~\cite{hu19randlanet, zhao21pt, park2022fast} and object detection~\cite{yang18pixor,zhou18voxelnet,qi19votenet,wang22brt}. 
A large portion of previous methods 
design different neural networks and learn from large-scale $annotated$ data for point cloud understanding tasks.
However, point cloud labels are rare in most scenarios and acquiring them is time-consuming and expensive. 

Hence, 
in recent years, researchers have begun to shift their attention to developing unsupervised methods for point cloud understanding,
without the need of hand-crafted annotations. 
Unsupervised methods are designed in different ways, such as auto-encoders~\cite{yang18foldingnet}, 
mask auto-encoders~\cite{yu2021pointbert, pang2022pointmae}, reconstruction~\cite{sauder19jigsaw}, 
occlusion completion~\cite{Wang_2021_ICCV, yu2021pointr}, GANs~\cite{wu16learning, achlioptas18learning, han19url} and 
contrastive learning~\cite{pointcontrast, depthcontrast, hou2021exploring, afham2022crosspoint, tang2022cbl}, etc.

\begin{figure}[t]
        \begin{center}
                \includegraphics[width=0.49\linewidth]{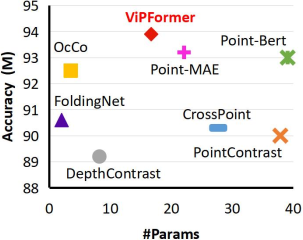}
                \includegraphics[width=0.49\linewidth]{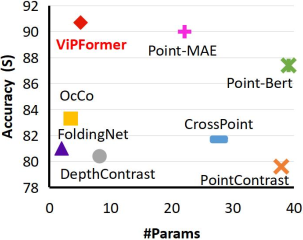}
        \end{center}
        \caption{Comparison of classification accuracy and the number of parameters of different 
        unsupervised methods on ModelNet40 (M) and ScanObjectNN (S), respectively.}
        \label{fig:method_acc_params}
\end{figure}

Currently, a growing number of methods embrace contrastive learning  
because it is a simple yet effective framework and has shown 
improvements in vision~\cite{he2022mae} and language processing~\cite{kim21vilt}. 
This framework can be easily extended to incorporate multi-modal data to further exploit richer semantics in
multi-modal data to improve performances. 
Inspired by the success of vision + language~\cite{desai21vertex, sariyildiz20learning} and video + audio~\cite{wwens18audio, morgado21audio}, point cloud understanding powered by cross-modal data 
has drawn research interests. 

CrossPoint~\cite{afham2022crosspoint} takes images and point clouds 
as inputs and follows the contrastive framework for unsupervised point cloud processing. It utilizes ResNet50~\cite{he2016resnet} as image feature extractor and 
PointNet~\cite{qi17pointnet}/DGCNN~\cite{wang19dgcnn} as point cloud feature extractor. 
\cite{jing22contrastive} uses CNN and U-Net~\cite{ronneberger15unet} architecture in the image branch and PointNet++~\cite{qi17pointnet2} architecture in the point cloud branch for contrastive learning. 
Although promising performances are obtained, the unbalanced image and point cloud 
processing architecture makes them unnecessarily complex and inefficient.
For example, in \cite{afham2022crosspoint}, the point cloud branch has 3M parameters while the image branch has 25M. 
The image processing branch is $\sim$8.3x heavier than the point cloud one and consumes much more time. 

The different and unbalanced architecture when dealing with data from different modalities
is often neglected in 
academic studies but is a critical problem in practice because it severely hinders efficiency. 
However, it is possible to design a unified and efficient architecture to process 
cross-modal data since Transformer~\cite{vaswani17transformer} has shown the flexibility and superiority in 
vision~\cite{dosovitskiy2021vit, bao2022beit, he2022mae} and 
language~\cite{devlin19bert, radford2019gpt2, brown20gpt3} modeling. 
And recently, Point-BERT~\cite{yu2021pointbert} and Point-MAE~\cite{pan21pointformer} show point clouds can be sampled to 
groups then processed by Transformer and the performances are promising.

In this paper, we propose an efficient \textbf{Vi}sion-and-\textbf{P}ointcloud Trans\textbf{former} (ViPFormer) for unsupervised 
point cloud understanding. 
ViPFormer unifies image and point cloud processing in a single architecture, which ensures the two branches have the same size and complexity. 
Then it follows the contrastive learning framework to optimize image and point cloud feature representations. 
Finally, the learned representations are transferred to target tasks like 3D point cloud classification and semantic segmentation. 

ViPFormer is evaluated on a wide range of point cloud understanding tasks, including 
synthetic and real-world 3D shape classification, object part segmentation. 
Results show it not only reduces the model complexity and running latency but also outperforms all existing unsupervised methods. 
The major contributions of this paper are summarized as follows:

\begin{itemize}
        \item We propose ViPFormer, handling image and point cloud data in a unified architecture, simplifies the model complexity, reduces
        running latency and boosts overall performances for unsupervised point cloud understanding. 
        \item We show that ViPFormer can be generalized better to different tasks by simultaneously optimizing intra-modal and cross-modal 
        contrastive objectives. 
        \item The proposed method is validated on various downstream tasks, e.g., 
        it achieves 90.7\% classification accuracy on ScanObjectNN, 
        leading CrossPoint by 9\%, and surpassing the previous best performing unsupervised method 
        by 0.7\% with $\sim$77\% fewer parameters.
        Similarly, ViPFormer reaches a 93.9\% score on ModelNet40, outperforming the previous state-of-the-art method and reducing the number of parameters by 24\%. 
        \item We conduct extensive ablation studies to clarify the advantages of 
        the architecture design, contrastive optimization objectives, and unsupervised
        learning strategy. 
\end{itemize}

\section{Related Work}
\label{sec:related_work}

\textbf{Unsupervised Point Cloud Understanding} 
Unsupervised learning becomes more and more popular since it can unleash the potential of large-scale unlabeled data and 
save considerable costs. 
According to the pretext task, unsupervised methods for point cloud understanding can be classified into generative models
and discriminative models. 
Generative models usually learn the latent representations of point clouds by predicting some parts or all of the input data. 
The assumption is that only after the model understands the point cloud can it predict the occluded parts or 
generate the entire point cloud.
Auto-encoders like FoldingNet~\cite{yang18foldingnet}, 
GANs like LRGM~\cite{achlioptas18learning}, URL~\cite{han19url} and 3D GAN~\cite{wu16learning}, 
reconstruction methods like JigSaw~\cite{sauder19jigsaw}, cTree~\cite{sharma20ctree}, 
all of them generate a whole point cloud and maximize the similarity with the input point cloud. 
Mask encoders like OcCo~\cite{Wang_2021_ICCV}, Point-BERT~\cite{yu2021pointbert}, Point-MAE~\cite{pang2022pointmae}
complete the masked parts of a point cloud to keep it same as the input. 
On the other hand, the discriminative models aim to learn discriminative features from different object/semantic categories.
Most of them follow the contrastive learning framework 
~\cite{pointcontrast, depthcontrast, hou2021exploring, nunes2022ral, tang2022cbl, afham2022crosspoint}, 
where CrossPoint~\cite{afham2022crosspoint} is the most relevant to us since 
it also fuses cross-modal data, images and point clouds, for point cloud understanding.
However, the unbalanced feature extractors in CrossPoint caused much higher running latency and model complexity. 
Instead, we propose Vision-and-Pointcloud Transformer to unify image and point cloud processing in a single architecture, 
reduce latency and boost performance. 

\textbf{Architecture for Image and Pointcloud Processing} 
An image consists of regular and dense pixel grids, while a point cloud is a set of irregular, sparse and unordered points. 
The huge differences make it difficult to process images and point clouds in the same way. 
Researchers developed different architectures for image and point cloud processing. 
In many cases, CNNs are the first choices of image processing and PointNet~\cite{qi17pointnet} and its variants~\cite{qi17pointnet2,ma2022rethinking} 
are good starts for 
point cloud processing. 
However, the situation has changed since the advent of Transformer~\cite{vaswani17transformer}. 
Due to the notable improvements, Transformer quickly became the de facto standard architecture for language 
understanding tasks~\cite{devlin19bert, radford2019gpt2, brown20gpt3} then entered vision~\cite{carion20detr, dosovitskiy2021vit, bao2022beit, he2022mae} and 
3D field~\cite{misra213detr, mao21votr, sheng21ctformer, liu21group, pan21pointformer}. 
Guo $et~al.$ and Zhao $et~al.$ proposed PCT~\cite{guo2020pct} and Point Transformer~\cite{zhao21pt}, respectively,
but their architectures were different from the standard Transformer~\cite{vaswani17transformer} and can not be generalized to vision modality. 
Perceiver~\cite{jaegle21perceiver} and PerceiverIO~\cite{jaegle22perceiverio} have taken important steps toward 
general architecture for various modalities (audio, image, point cloud). However, Perceiver and PerceiverIO learn in a supervised fashion. 
Differently, the proposed ViPFormer unifies image and point cloud processing in a single architecture and 
learns from large-scale unlabeled data. 

\begin{figure*}[t]
        \begin{center}
                \includegraphics[width=0.9\linewidth]{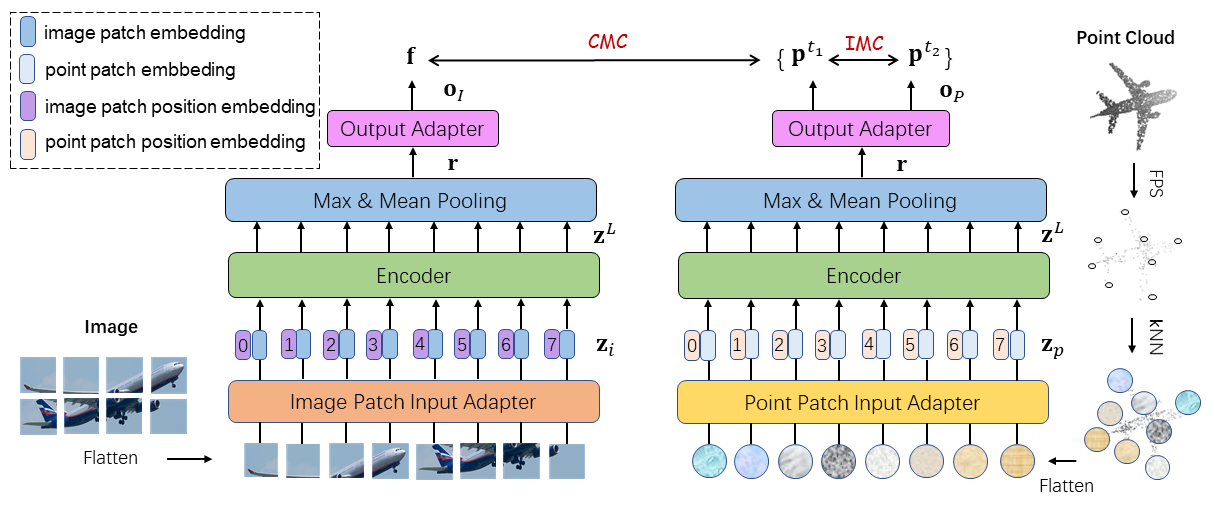}
        \end{center}
        \caption{The overall architecture of ViPFormer.}
        \label{fig:architecture}
\end{figure*}

\section{Methodology}
\label{sec:methodology}
In this section,  
firstly, we introduce the overall architecture of ViPFormer. 
Secondly, we elaborate on its unsupervised learning strategy.

\subsection{The Overall Architecture of ViPFormer}
\label{ssec:architecture}
As Fig.~\ref{fig:architecture} shows, ViPFormer consists of three components, which are a lightweight Input Adapter, 
Transformer Encoder and Output Adapter. In image and point cloud branches, 
modules with the same color are identical. And the images
and point clouds are serialized in different ways. 

\textbf{Image and Point Cloud Preparing} 
To exploit the power of Transformer~\cite{vaswani17transformer} 
we need to convert images and point clouds into sequence data as Transformer requires. 
Inspired by ViT~\cite{dosovitskiy2021vit}, we divide an image into small patches and then flatten them into a sequence. 
For example, an image $I$ is of size $H\times W\times C_1$ and the patch size is $Q$, we can generate an image patch sequence 
$\textbf{x}_i \in \mathbb{R}^{M\times(Q^2\cdot C_1)}$, where $M = HW/Q^2$.

For a point cloud $P$ of size $N\times C_2$, we convert it into a patch sequence as follows ~\cite{yu2021pointbert, pang2022pointmae}. First, the 
farthest point sampling (FPS) is applied to $P$ to get $G$ centers. Second, for each center, we search 
its $k$ nearest neighbors ($k$NN) in $P$ to aggregate local geometry information, resulting in a patch sequence 
$\textbf{x}_p \in \mathbb{R}^{G\times (k\cdot C_2)}$.

\textbf{Input Adapter} We design a lightweight image patch adapter \textbf{E}$_{I}$ and a point patch adapter \textbf{E}$_{P}$ to project 
the sequences to high dimensional feature representations. 
\textbf{E}$_{I}$ $\in \mathbb{R}^{(Q^2\cdot C_1)\times D}$ is a linear layer and \textbf{E}$_{P}$ is a multi-layer perception (MLP). 
The outputs are denoted as image patch embeddings \textbf{z}$_i$ and point patch embeddings \textbf{z}$_p$. 
\begin{equation}
        \textbf{z}_i = \textbf{x}_i \textbf{E}_{I}, \quad
        \textbf{z}_p = \textbf{x}_p \textbf{E}_{P}
\end{equation}
Before being fed into Encoder, the position information is injected to \textbf{z}$_i$ and \textbf{z}$_p$ by adding the image patch position 
embeddings $\textbf{E}_{I}^{pos}$ and point patch position embeddings $\textbf{E}_{P}^{pos}$. 
\begin{equation}
        \textbf{z}_i = \textbf{z}_i + \textbf{E}_{I}^{pos}, \quad
        \textbf{z}_p = \textbf{z}_p + \textbf{E}_{P}^{pos}
\end{equation}

\textbf{Encoder} 
The image and point cloud branches share the Encoder architecture, which 
ensures image and point cloud processing have balanced complexity and low latency. 
ViPFormer Encoder consists of $L$ stacked multi-head self-attention (MSA) and MLP layers. 
MLP has two layers with a GELU non-linear activation.
LayerNorm (LN) is applied before MSA and MLP layers, while Dropout is applied after them. 
\begin{equation}
        \hat{\textbf{z}}^{l} = \textrm{Dropout}(\textrm{MSA}(\textrm{LN}(\textbf{z}^{l-1}))) + \textbf{z}^{l-1}, \quad l=1 ... L
\end{equation}
\begin{equation}
        \textbf{z}^{l} = \textrm{Dropout}(\textrm{MLP}(\textrm{LN}(\hat{\textbf{z}}^{l}))) + \hat{\textbf{z}}^{l}, \quad l=1 ... L
\end{equation}
Before proceeding, the output sequence of ViPFormer Encoder needs to be converted into an object-level feature. 
We implement it by concatenating the max and mean value of \textbf{z}$^L$.
\begin{equation}
        \textbf{r} = \textrm{Concat}(\textrm{Max}(\textbf{z}^L), \textrm{Mean}(\textbf{z}^L))
\end{equation}

\textbf{Output Adapter} The image and point cloud branches also share the Output Adapter. 
As suggested by SimCLR~\cite{chen20simclr}, a learnable nonlinear transformation 
between the encoder and the contrastive loss can improve the quality of feature representations. 
The Output Adapter is implemented by two consecutive Linear layers, 
preceding with BatchNorm (BN) and ReLU. 

\begin{equation}
        \hat{\textbf{r}} = \textrm{Linear}(\textrm{ReLU}(\textrm{BN}(\textbf{r})))
\end{equation}
\begin{equation}
        \textbf{o} = \textrm{Linear}(\textrm{ReLU}(\textrm{BN}(\hat{\textbf{r}})))
\end{equation}

At this point, the input image $I$ and point cloud $P$ are transformed into image feature $\textbf{f}$ = \textbf{o}$_I$ and point cloud feature
$\textbf{p}$ = \textbf{o}$_P$. We can use these features for unsupervised contrastive learning. 

\subsection{Unsupervised Contrastive Pretraining of ViPFormer}
\label{ssec:optimization}
We conduct unsupervised pretraining for ViPFormer by introducing two contrastive objectives, 
intra-modal contrast and cross-modal contrast. They are formulated as follows.

\textbf{The Intra-Modal Contrastive (IMC) Objective}
injects ViPFormer with the ability to resist data transformations and small perturbations (e.g., translation, rotation, jittering) 
to the same objects while maximizing the distance of different objects in feature space. 
This strategy will make the pretrained model insensitive to random noises and generalize better. 
Specifically, a point cloud $P$ is transformed by two data augmentations $t_1$ and $t_2$, resulting in $P^{t_1}$ and $P^{t_2}$. 
After going through ViPFormer, their feature representations 
$\textbf{p}^{t_1}$=$\textbf{o}_{P}^{t_1}$ and $\textbf{p}^{t_2}$=$\textbf{o}_{P}^{t_2}$ are obtained. 
The IMC objective $\mathcal{L}_{imc}$ is formulated by NT-Xent loss~\cite{chen20simclr}: 
\begin{equation}
        l(i,\!t_1,\!t_2)\!\!=\!\!-\!\log\frac{\exp(s(\textbf{p}_{i}^{t_1},\textbf{p}_{i}^{t_2})/\tau\!)}
        {\sum\limits_{k=1\atop k\neq i}^{N}\!\!\exp(s(\textbf{p}_{i}^{t_1}\!,\!\textbf{p}_{k}^{t_1})/\tau)\!\!+\!\!\sum\limits_{k=1}^{N}\!\!\exp(s(\textbf{p}_{i}^{t_1}\!,\!\textbf{p}_{k}^{t_2})/\tau)}
        \label{eq:imc1}
\end{equation}
\begin{equation}
        \mathcal{L}_{imc}  = \frac{1}{2N} \sum_{i=1}^{N} (l(i, t_1, t_2) + l(i, t_2, t_1))
        \label{eq:imc2}
\end{equation}
where $N$ is the batch size, $\tau$ is the temperature coefficient and $s(\cdot)$ represents the cosine similarity. 

\textbf{The Cross-Modal Contrastive (CMC) Objective}
maximizes the agreement of feature representations of paired images and point clouds, 
while minimizing that of unpaired ones in the same feature space. ViPFormer achieves this goal
only when it understands the information contained in both modalities. Similarly, the CMC objective $\mathcal{L}_{cmc}$ 
is formulated by NT-Xent loss~\cite{chen20simclr}:
\begin{equation}
        l(i,\textbf{p},\textbf{f})\!=\!-\!\log\frac{\exp(s(\textbf{p}_i,\textbf{f}_i)/\tau)}{\sum\limits_{k=1\atop k\neq i}^{N}\!\exp(s(\textbf{p}_i,\textbf{p}_k)/\tau)\!\!+\!\!\sum\limits_{k=1}^{N}\!\exp(s(\textbf{p}_i,\textbf{f}_k)/\tau)}
        \label{eq:cmc1}
\end{equation}
\begin{equation}
        \mathcal{L}_{cmc}  = \frac{1}{2N} \sum_{i=1}^{N} (l(i,\textbf{p},\textbf{f}) + l(i,\textbf{f},\textbf{p}))
        \label{eq:cmc2}
\end{equation}
$N$, $\tau$ and $s(\cdot)$ have the same meaning as those in Eq.~\ref{eq:imc1}. 

During pretraining, ViPFormer combines IMC and CMC as the overall loss. 
A balancing factor $\alpha$ is deployed between two objectives as the cross-modal loss $\mathcal{L}_{cmc}$ is harder to optimize and usually several times bigger 
than the intra-modal loss $\mathcal{L}_{imc}$. 
\begin{equation}
        \mathcal{L} = \mathcal{L}_{imc} + \alpha \mathcal{L}_{cmc}
\end{equation}

\section{Experiments}
\label{sec:experiments}

In this section, firstly, we elaborate on the pretraining settings of ViPFormer. 
Then the pretrained ViPFormer is transferred to various downstream tasks to evaluate its performances. 
Thirdly, the effectiveness of different components in ViPFormer is validated by extensive ablation studies. 
Finally, the predictions of ViPFormer on different tasks are visualized for a better understanding. 

\subsection{Pretraining Settings}
\textbf{Datasets}
We use the same dataset as in~\cite{afham2022crosspoint}. The 
point clouds and images come from ShapeNet~\cite{shapenet2015} and DISN~\cite{xu19disn}. 
There are 43,783 point clouds and each point cloud $P$ corresponds to 24 rendered images, from which 
an image $I$ is randomly selected to pair with $P$. 
During pretraining, point cloud $P$ contains 2048 points and the corresponding image $I$ is resized to 144$\times$144$\times$3. 
In FPS and $k$NN, the point cloud is divided into $G$=128 centers and $k$=32 nearest neighbors are retrieved for each center.
The image patch size $Q$ is set to 12. 

\textbf{Architecture} 
In Input Adapter, the dimension of point/image patch embedding is projected to 384.
In Encoder, there are $L$=9 stacked MSA and MLP layers. All MSA layers have 6 heads. 
The widening ratio of the MLP hidden layer is 4. 
In Output Adapter, the 2-layer MLP is of size \{768, 384, 384\}. 
We justify the design choices of the architecture through controlled 
experiments in Section~\ref{sssec:architecture}.

\textbf{Optimization} 
We pretrain ViPFormer for 300 epochs, adopting AdamW~\cite{loshchilov2018decoupled} as the optimizer and 
CosineAnnealingWarmupRestarts~\cite{Katsura21cawwr} as the learning rate scheduler.
The restart interval is 100 epochs and the warmup period is the first 5 epochs. 
The learning rate scales linearly to peak during each warmup, then decays with the cosine annealing schedule. 
The initial learning rate peak is 0.001, multiplied by 0.6 after each interval. The balancing factor $\alpha$ 
is set to 1, which works well. 
We record the best pretrained model according to the zero-shot accuracy on ModelNet40~\cite{wu15modelnet}.

\subsection{Model Complexity, Latency and Performance on Downstream Tasks} 
In this part, the pretrained ViPFormer is transferred to various downstream tasks to evaluate its complexity, 
latency and performance. 
These metrics are critical dimensions for assessing point cloud understanding methods. 
Complexity is reflected by a model's number of parameters (\#Params). 
Latency is counted by running time and performance is subject to overall accuracy (OA) in the classification task
and mean class-wise Intersection of Union (mIoU) in the segmentation task.
We compare with previous state-of-the-art unsupervised methods. 

\textbf{Point Cloud Classification} 
The experiments  are conducted on two widely used datasets: 
ScanObjectNN~\cite{Uy_2019_ICCV} and ModelNet40~\cite{wu15modelnet}. 
ScanObjectNN contains 2880 objects from 15 categories. 
It is challenging because objects in this dataset are usually cluttered with background or are partial due to occlusions. 
ModelNet40 is a synthetic point cloud dataset, including 12308 objects across 40 categories. 
We use the same settings as previous work~\cite{pointcontrast,depthcontrast,Wang_2021_ICCV,afham2022crosspoint} to sample 1024 points to represent a 3D object.
We reimplement previous methods according to the released codes since they do not report the \#Params and latency. 
For latency, we consider two stages (Pretrain and Finetune) and count the running time of a single epoch in each stage. 

For the ScanObjectNN~\cite{Uy_2019_ICCV} dataset, all methods are finetuned on it and 
the results are recorded in Tab.~\ref{tab:different_stage_efficiency}. The best score is in bold black and the second best score 
is marked in blue. 
ViPFormer not only outperforms previous state-of-the-art Point-MAE by 0.7\% in classification accuracy but also reduces
76.9\% \#Params and runs $\sim$2.6x faster than Point-MAE. 

\begin{table}[ht]
        \centering
        \caption{Comparison of model complexity, latency and performance with existing unsupervised methods
        on ScanObjectNN.}
        \begin{tabular}{l r r r r }
                \toprule
                \multirow{3}{*}{Method} & \#Params & \multicolumn{2}{c}{Latency} & OA \\  
                &  & Pretrain & Finetune & \\
                & (M) & (s) & (ms) & (\%) \\
                \midrule
                FoldingNet~\cite{yang18foldingnet} & \textbf{2.0} & -- & -- & 81.0 \\
                PointContrast~\cite{pointcontrast} & 37.9 & -- & -- & 79.6 \\
                DepthContrast~\cite{depthcontrast} & 8.2 & -- & -- & 80.4 \\
                OcCo~\cite{Wang_2021_ICCV} & \textcolor{blue}{3.5} & $\sim$600.0 & 16,100  & {83.3} \\
                CrossPoint~\cite{afham2022crosspoint} & 27.7 & 946.0 & 14,000 & {81.7} \\
                Point-BERT~\cite{yu2021pointbert} & 39.1 & 633.5 & 3,973 & {87.4} \\
                Point-MAE~\cite{pang2022pointmae} & 22.1 & \textcolor{blue}{576.0} & \textcolor{blue}{3,612} & \textcolor{blue}{90.0} \\
                \midrule
                \textbf{ViPFormer} & 5.1 & \textbf{22.2} & \textbf{1,015} & \textbf{90.7} \\ 
                \bottomrule
        \end{tabular}
        \label{tab:different_stage_efficiency}
\end{table}

The classification results on ModelNet40 are shown in Table~\ref{tab:cls_params_flops_accuracy}. 
The Pretrain latency is not changed because pretraining is independent of the downstream datasets, 
including ScanObjectNN and ModelNet40. 
ModelNet40 is a larger dataset so finetuning on it consumes more time. 
ViPFormer achieves higher classification accuracy with lower model complexity and runtime latency. 
It leads Point-MAE by 0.7\% accuracy while reducing \#Params by 24.1\%.

\begin{table}[ht]
        \centering
        \caption{Comparison of model complexity, latency and performance with existing unsupervised methods
        on ModelNet40.}
        \begin{tabular}{l r r r r }
                \toprule
                \multirow{3}{*}{Method} & \#Params & \multicolumn{2}{c}{Latency} & OA \\
                &  & Pretrain & Finetune & \\
                & (M) & (s) & (ms) & (\%) \\
                \midrule
                FoldingNet~\cite{yang18foldingnet} & \textbf{2.0} & -- & -- & 90.6 \\    
                PointContrast~\cite{pointcontrast} & 37.9 & -- & -- & 90.0 \\ 
                DepthContrast~\cite{depthcontrast} & 8.2 & -- & -- & 89.2 \\ 
                OcCo~\cite{Wang_2021_ICCV} & \textcolor{blue}{3.5} & $\sim$600.0 &  39,295 & 92.5 \\        
                CrossPoint~\cite{afham2022crosspoint} & 27.7 & 946.0 & 35,258 & 90.3 \\       
                Point-BERT~\cite{yu2021pointbert} & 39.1 & 633.5 & 10,329 & 93.0 \\        
                Point-MAE~\cite{pang2022pointmae} & 22.1 & \textcolor{blue}{576.0} & \textcolor{blue}{9,344} & \textcolor{blue}{93.2} \\
                \midrule
                \textbf{ViPFormer} & 16.7  & \textbf{60.9} & \textbf{4,198} & \textbf{93.9} \\         
                \bottomrule
        \end{tabular}
        \label{tab:cls_params_flops_accuracy}
\end{table}

\textbf{Object Part Segmentation}
We also transfer ViPFormer to the task of object part segmentation. The experiments are conducted on the 
ShapeNetPart~\cite{shapenetpart} dataset which contains 16881 point clouds 
and each point cloud  consists of 2048 points. 
Objects in ShapeNetPart are divided into 16 categories and 50 annotated parts. 
For a fair comparison, we follow previous work~\cite{yu2021pointbert}~\cite{pang2022pointmae} to add a
simple part segmentation head on ViPFormer Encoder. The pretrained weights of ViPFormer are used to
initialize the part segmentation model. 
In addition to the above metrics, mean class-wise IoU (mIoU) is added to evaluate the part segmentation performance. 
The results are reported in Tab.~\ref{tab:partseg_params_latency_mIoU}.
ViPFormer reaches comparable OA and mIoU with best performing Point-MAE while having lower 
model complexity. 

\begin{table}[ht]
        \centering
        \caption{Object part segmentation on ShapeNetPart.}
        \begin{tabular}{l r r r r}
                \toprule
                \multirow{2}{*}{Method} & \#Params  & Latency & OA & mIoU \\    
                & (M) & (s) & (\%) & (\%)\\
                \midrule
                PointContrast~\cite{pointcontrast} & 37.9 & -- & -- & -- \\    
                OcCo~\cite{Wang_2021_ICCV} & \textbf{1.5}  & \textbf{32.0} & 93.9 & 79.7 \\    
                CrossPoint~\cite{afham2022crosspoint} & 27.5 & 80.0 & 93.8 & 84.3 \\ 
                Point-BERT~\cite{yu2021pointbert} & 44.1  & 58.8 & -- & 84.1 \\    
                Point-MAE~\cite{pang2022pointmae} & 27.1  & 46.3 & \textcolor{blue}{94.8} & \textcolor{blue}{84.7} \\
                \midrule
                \textbf{ViPFormer} & \textcolor{blue}{26.8}  & \textcolor{blue}{42.1} & \textbf{94.8} & \textbf{84.7} \\
                \bottomrule
        \end{tabular}
        \label{tab:partseg_params_latency_mIoU}
\end{table}

\textbf{Few-shot Object Classification} Few-shot evaluation is used to validate the transferability of a pretrained model with limited labeled data. 
The conventional setting is ``$N$-way, $K$-shot''~\cite{depthcontrast,Wang_2021_ICCV,afham2022crosspoint}. 
Under this setting, $N$ classes and $K$ samples in a downstream task dataset are randomly selected for training an SVM model of the linear kernel. 
The test score on the downstream task given by SVM can reflect the quality of the pretrained model as
the inputs to the SVM model are the features extracted by the pretrained model. 
Here the downstream task datasets are ModelNet40 and ScanObjectNN, respectively. 
We perform 10 runs for each ``$N$-way, $K$-shot'' and compute their mean and standard deviation. 
The results are shown in Table~\ref{tab:fewshot_cls}. On ModelNet40, 
ViPFormer achieves comparable accuracy with previous strong baselines, whereas it shows 
consistent improvements on ScanObjectNN. 
The IMC and CMC objectives enable ViPFormer to understand the information contained in both modalities, 
so it can better deal with the challenging scenarios in ScanObjectNN.

\begin{table}[t]
        \centering
        \caption{Comparison of few-shot classification accuracy with existing methods on ModelNet40 and ScanObjectNN.}
        \begin{tabular}{l | c c | c c}
                \toprule
                \multirow{2}{*}{Method} & \multicolumn{2}{c|}{5-way} & \multicolumn{2}{c}{10-way} \\\cline{2-5}
                                        & 10-shot & 20-shot & 10-shot & 20-shot \\
                \midrule
                & \multicolumn{4}{c}{ModelNet40} \\
                OcCo~\cite{Wang_2021_ICCV} & 90.6$\pm$2.8 & 92.5$\pm$1.9 & \textbf{82.9$\pm$1.3} & 86.5$\pm$2.2\\      
                CrossPoint~\cite{afham2022crosspoint} & \textcolor{blue}{91.0$\pm$2.9} & \textbf{95.0$\pm$3.4} & \textcolor{blue}{82.2$\pm$6.5} & \textbf{87.8$\pm$3.0}\\        
                \textbf{ViPFormer} & \textbf{91.1$\pm$7.2} & \textcolor{blue}{93.4$\pm$4.5} & 80.8$\pm$4.2 & \textcolor{blue}{87.1$\pm$5.8}\\
                 \midrule
                & \multicolumn{4}{c}{ScanObjectNN} \\
                OcCo~\cite{Wang_2021_ICCV} & 72.4$\pm$1.4 & 77.2$\pm$1.4 & 57.0$\pm$1.3 & 61.6$\pm$1.2\\      
                CrossPoint~\cite{afham2022crosspoint} & \textcolor{blue}{72.5$\pm$8.3} & \textcolor{blue}{79.0$\pm$1.2} & \textcolor{blue}{59.4$\pm$4.0} & \textcolor{blue}{67.8$\pm$4.4}\\        
                \textbf{ViPFormer} & \textbf{74.2$\pm$7.0} & \textbf{82.2$\pm$4.9} & \textbf{63.5$\pm$3.8} & \textbf{70.9$\pm$3.7}\\
                \bottomrule
        \end{tabular}
        \label{tab:fewshot_cls}
\end{table}

\subsection{Ablation Studies} 
\label{subsec:ablation}

Ablation studies are conducted to 1) justify the architecture of ViPFormer,  
2) demonstrate the effectiveness of IMC and CMC optimization objectives, 
and 3) analyze the advantages of the pretrain-finetune strategy over training from scratch. 

\subsubsection{Architecture} 
\label{sssec:architecture}

The controlled variables of ViPFormer architecture are 
the number of self-attention layers (\#SA\_Layers), 
the widening ratio of the MLP hidden layer (MLP ratio), 
the number of attention heads (\#Heads), 
the sequence length (\#Length) and the model dimension (D\_{model}). 
For different architectures, the accuracy of the pretrain-finetune scheme is reported on ModelNet40 and ScanObjectNN, 
respectively, shown in Tab.~\ref{tab:ablation_architecture}.
The overall trend is the larger the model, the better the performance. 
We choose the best-performing architecture to compare with other methods. 

\begin{table*}[t]
        \centering
        \caption{Ablation Study: Model Architecture.}
        \begin{tabular}{l r r r r r r r r r r r r r r r r r}
                \toprule
                \#SA\_Layers & 7 & 7 & 9 & 9 & 7 & 7 & 9 & 9 &  7 & 7 & 9 & 9 & 7 & 7 & 9 & 9 \\ 
                MLP ratio & 2 & 2 & 2 & 2 & 4 & 4 & 4 & 4 & 2 & 2 & 2 & 2 & 4 & 4 & 4 & 4 \\
                \#Heads & 4 & 4 & 4 & 4 & 4 & 4 & 4 & 4 & 6 & 6 & 6 & 6 & 6 & 6 & 6 & 6 \\
                \#Length & 96 & 128 & 96 & 128 & 96 & 128 & 96 & 128 & 96 & 128 & 96 & 128 & 96 & 128 & 96 & 128 \\
                D\_{model} & 256 & 256 & 256 & 256 & 256 & 256 & 256 & 256 & 384 & 384 & 384 & 384 & 384 & 384 & 384 & 384 \\
                \midrule
                Accuracy \\
                \quad ModelNet40 & 91.5 & 91.5 & 91.5 & 92.5 & 93.2 & 91.0 & 92.7 & 91.3 & 93.0 & 93.2 & 92.0 & 92.2 & 92.2 & 92.2 & \textcolor{blue}{93.2} & \textbf{93.9} \\
                \quad ScanObjectNN & 84.5 & 83.5 & 86.6 & \textbf{90.7} & 85.6 & 85.6 & 86.6 & 87.6 & 88.7 & 84.5 & 84.5 & 89.7 & 89.7 & 88.7 & 89.7 & \textcolor{blue}{89.7} \\
                \bottomrule
        \end{tabular}
        \label{tab:ablation_architecture}
\end{table*}

\subsubsection{Contrastive Optimization Objectives} 
\label{sssec:contrast_objectives}

The effectiveness of proposed IMC and CMC contrastive objectives are evaluated by training ViPFormer 
in three modes: i) only use IMC, ii) only use CMC, and iii) use IMC and CMC together. 
The experiments are conducted on different learning stages (Pretrain vs. Finetune) and different datasets (ModelNet40 vs. ScanObjectNN). 
The results are shown in Tab.~\ref{tab:ablation_modality}. 
Apparently, the combination of IMC and CMC optimization objectives significantly improves the model performance for 
target tasks across different datasets. 

\begin{table}[t]
        \centering
        \caption{Ablation Study: Performance comparison on ModelNet40 (M) and ScanObjectNN (S) when using different contrastive objectives.}
        \begin{tabular}{l | c c | c c}
                \toprule
                \multirow{2}{*}{Modality Types} & \multicolumn{2}{c|}{Pretrain} & \multicolumn{2}{c}{Finetune} \\
                 & OA$_{\mathcal{M}}$ & OA$_{\mathcal{S}}$ & OA$_{\mathcal{M}}$ & OA$_{\mathcal{S}}$ \\
                \midrule
                IMC $only$ & 86.4 & \textbf{76.4} & 91.3 & \textcolor{blue}{87.6} \\
                CMC $only$ & \textbf{87.3} & 66.4 & \textcolor{blue}{91.3} & 81.4 \\
                IMC \& CMC & \textcolor{blue}{87.0} & \textcolor{blue}{75.7} & \textbf{93.9} & \textbf{89.7} \\
                \bottomrule
        \end{tabular}
        \label{tab:ablation_modality}
\end{table}

\subsubsection{Learning Strategies}
\label{sssec:pt_ft_vs_scratch}

The differences between two kinds of learning strategies, Train from scratch and Pretrain-Finetune, are also investigated. 
As Tab.~\ref{tab:ablation_pt_ft_scratch} shows, 
The Pretrain-Finetune strategy outperforms Train from scratch by 1.9\% and 4.1\% on ModelNet40 and ScanObjectNN, respectively. 
The results indicate the initialization provided by the pretrained ViPFormer really helps the model find better 
directions and solutions in downstream tasks. 

\begin{table}[t]
        \centering
        \caption{Ablation Study: Comparison of the classification accuracy using Pretrain-Finetune and Train-from-scratch strategy. 
        The used datasets are ModelNet40 (M) and ScanObjectNN (S).}
        \begin{tabular}{l | c c }
                \toprule
                Learning Strategy & OA$_{\mathcal{M}}$ & OA$_{\mathcal{S}}$ \\   
                \midrule
                Train from scratch & 92.0 & 85.6 \\    
                Pretrain-Finetune & \textbf{93.9} & \textbf{89.7} \\    
                \bottomrule
        \end{tabular}
        \label{tab:ablation_pt_ft_scratch}
\end{table}

\subsection{Visualization} 

\begin{figure}[H]
        \centering
        \begin{subfigure}{0.24\linewidth}
          \includegraphics[width=\linewidth]{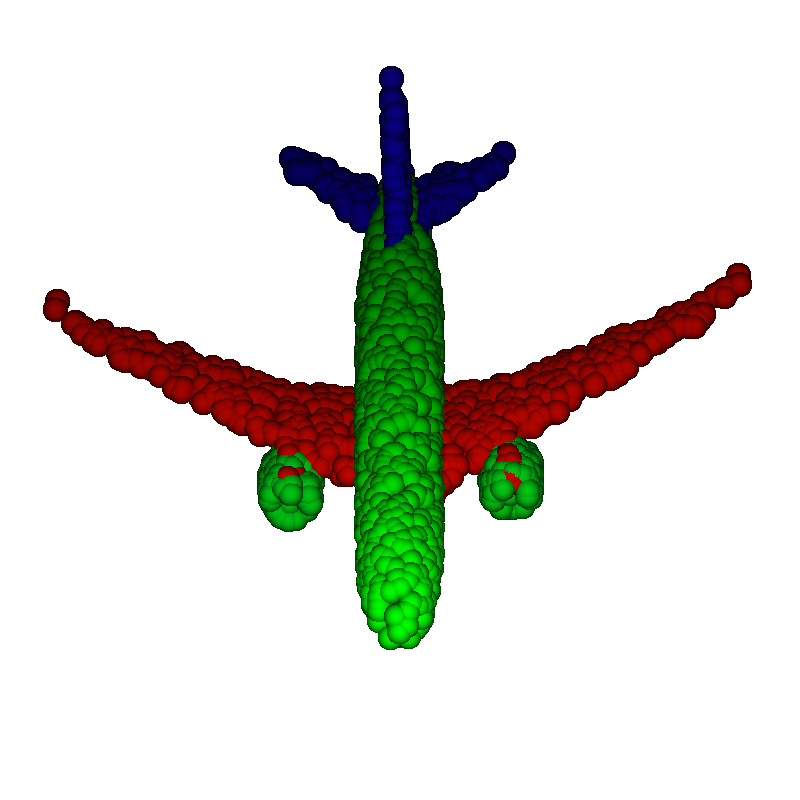}
          \caption{Airplane}
          \label{fig:pred_Airplane}
        \end{subfigure}
        \begin{subfigure}{0.24\linewidth}
           \includegraphics[width=\linewidth]{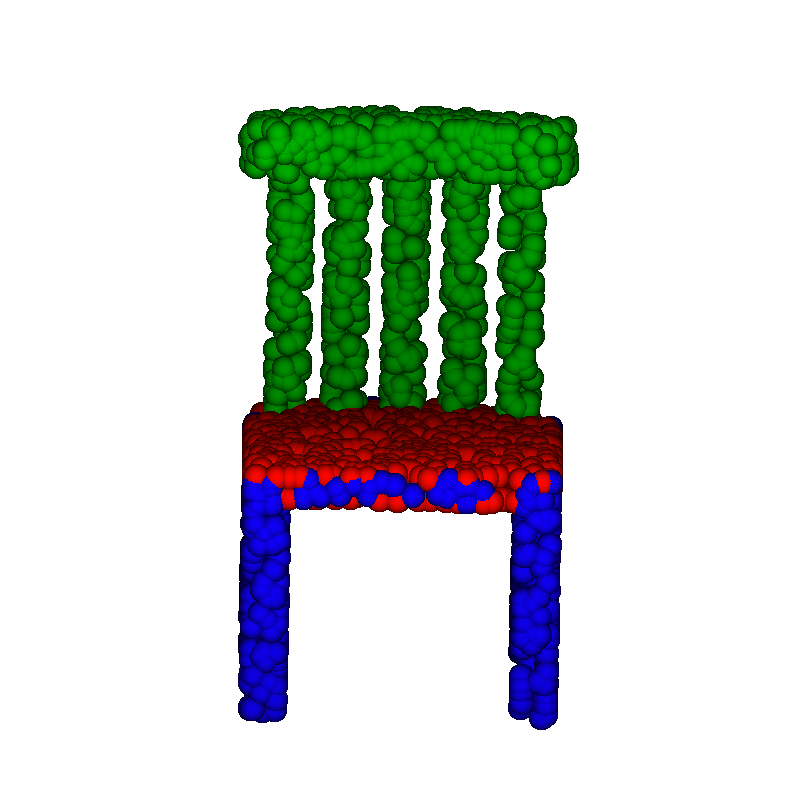}
           \caption{Chair}
           \label{fig:pred_Chair}
        \end{subfigure}
        \begin{subfigure}{0.24\linewidth}
                \includegraphics[width=\linewidth]{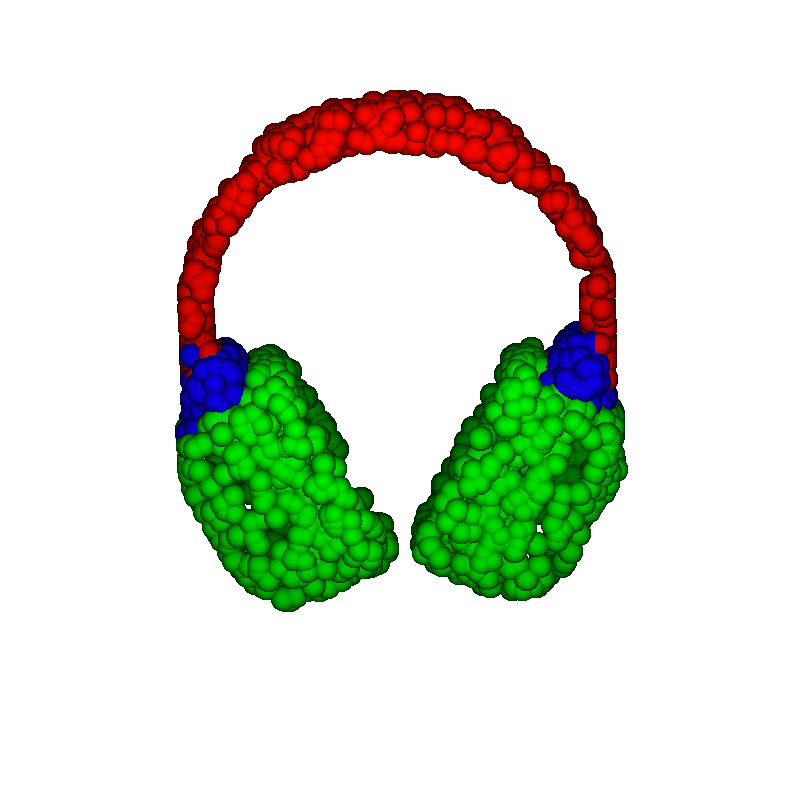}
                \caption{Earphone}
                \label{fig:pred_Earphone}
        \end{subfigure}
        \begin{subfigure}{0.24\linewidth}
                \includegraphics[width=\linewidth]{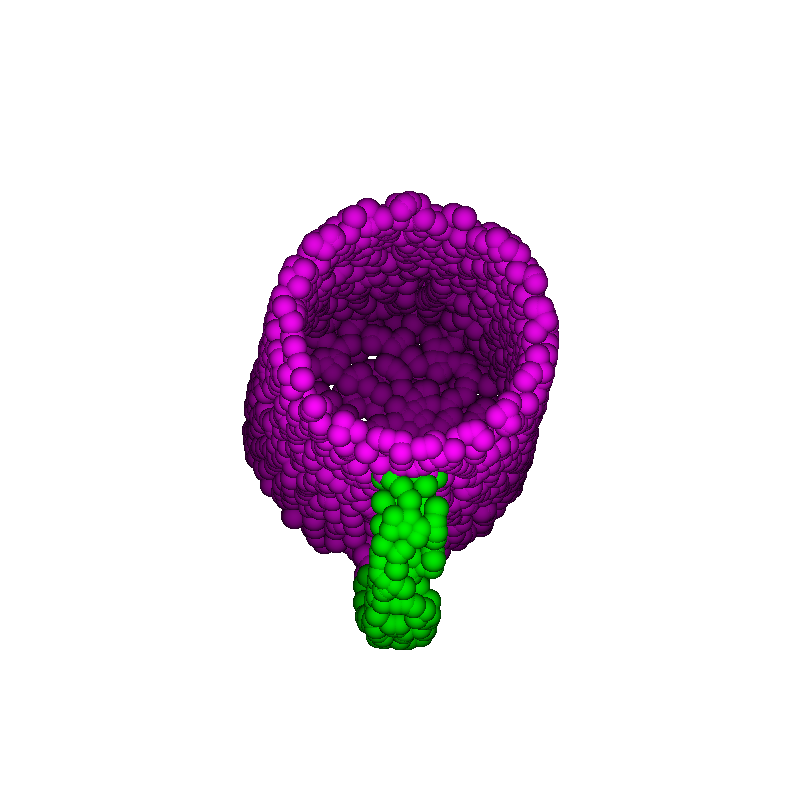}
                \caption{Mug}
                \label{fig:pred_Mug}
        \end{subfigure}
        
        \begin{subfigure}{0.24\linewidth}
                \includegraphics[width=\linewidth]{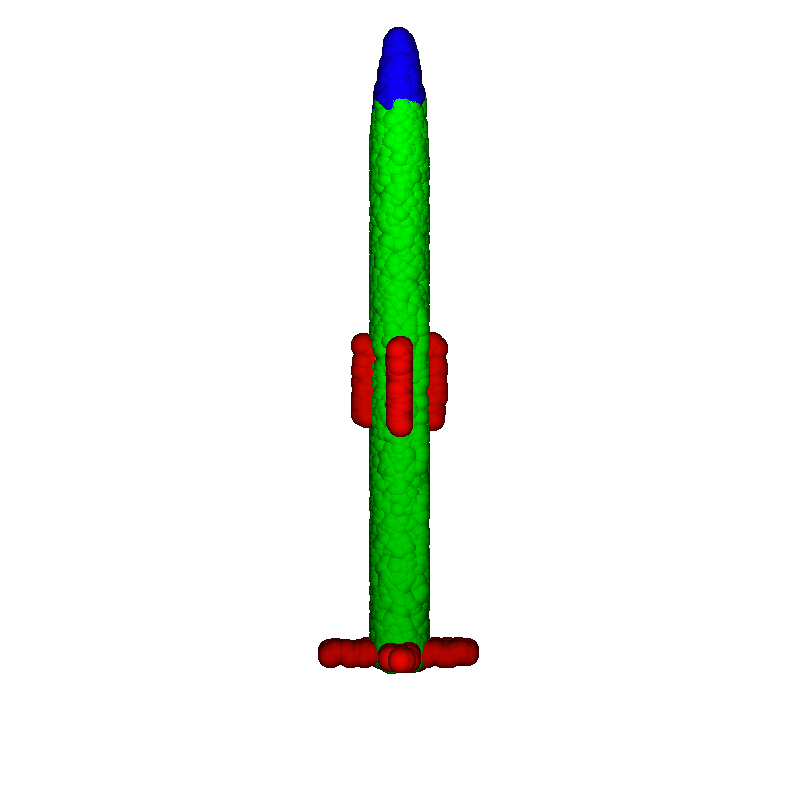}
                \caption{Rocket}
                \label{fig:pred_Rocket}
        \end{subfigure}
        \begin{subfigure}{0.24\linewidth}
                \includegraphics[width=\linewidth]{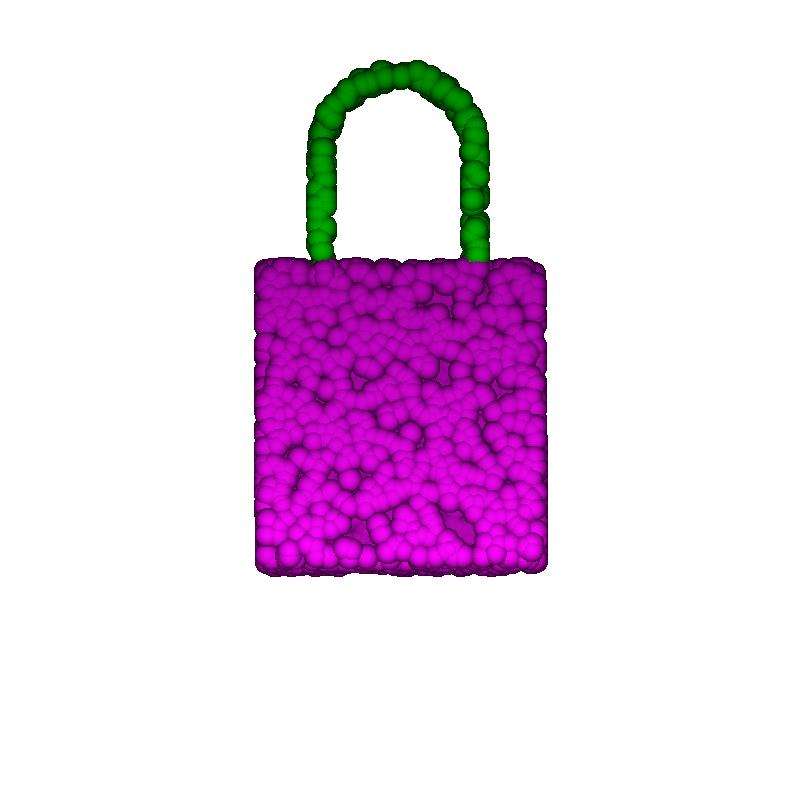}
                \caption{Bag}
                \label{fig:pred_Bag}
              \end{subfigure}
        \begin{subfigure}{0.24\linewidth}
                \includegraphics[width=\linewidth]{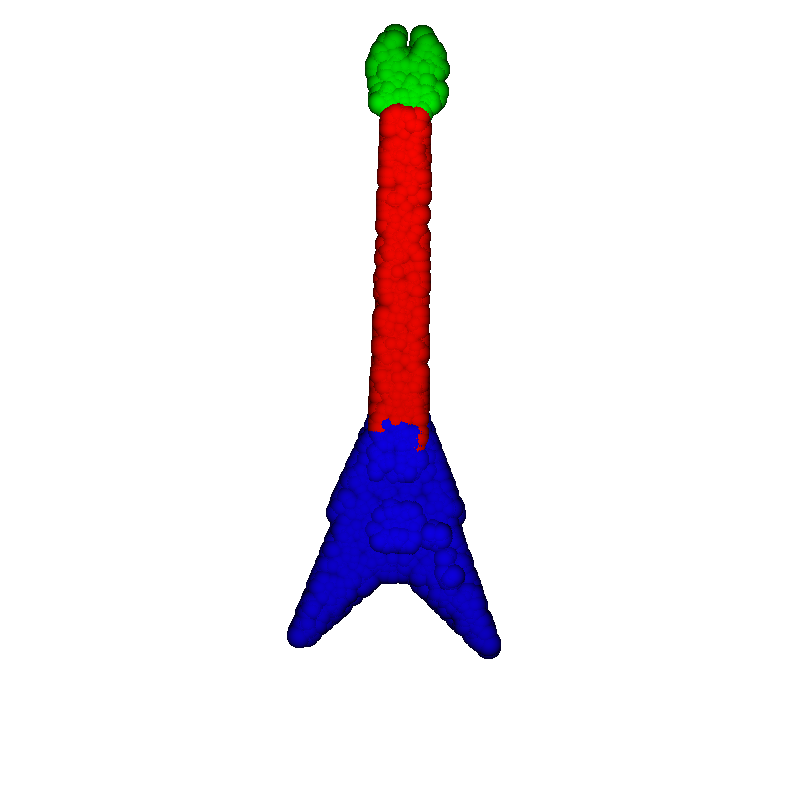}
                \caption{Guitar}
                \label{fig:pred_Guitar}
        \end{subfigure}
        \begin{subfigure}{0.24\linewidth}
                \includegraphics[width=\linewidth]{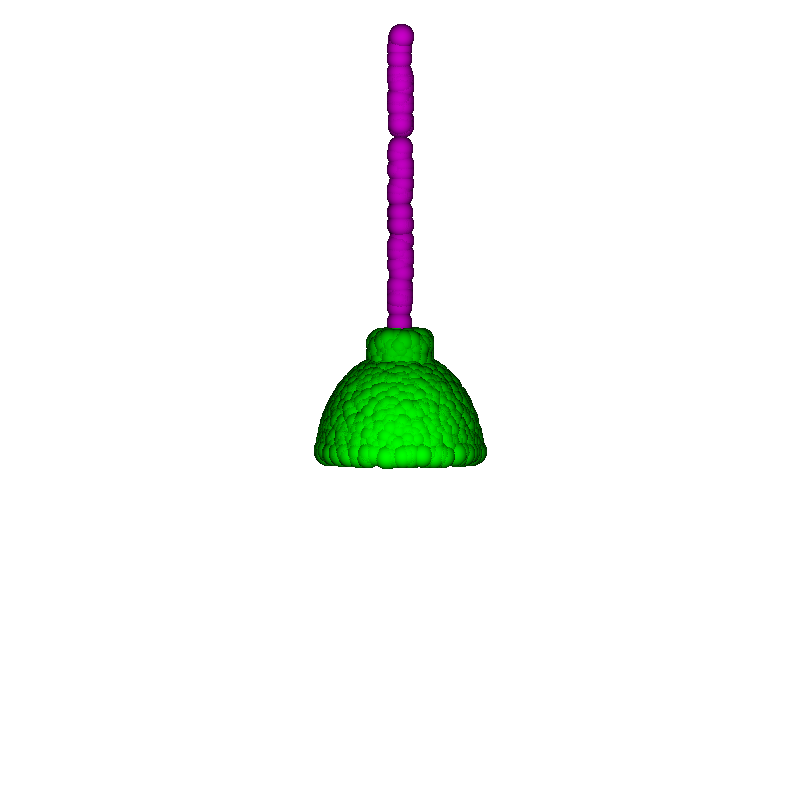}
                \caption{Lamp}
                \label{fig:pred_Lamp}
        \end{subfigure}
        \begin{subfigure}{0.24\linewidth}
                \includegraphics[width=\linewidth]{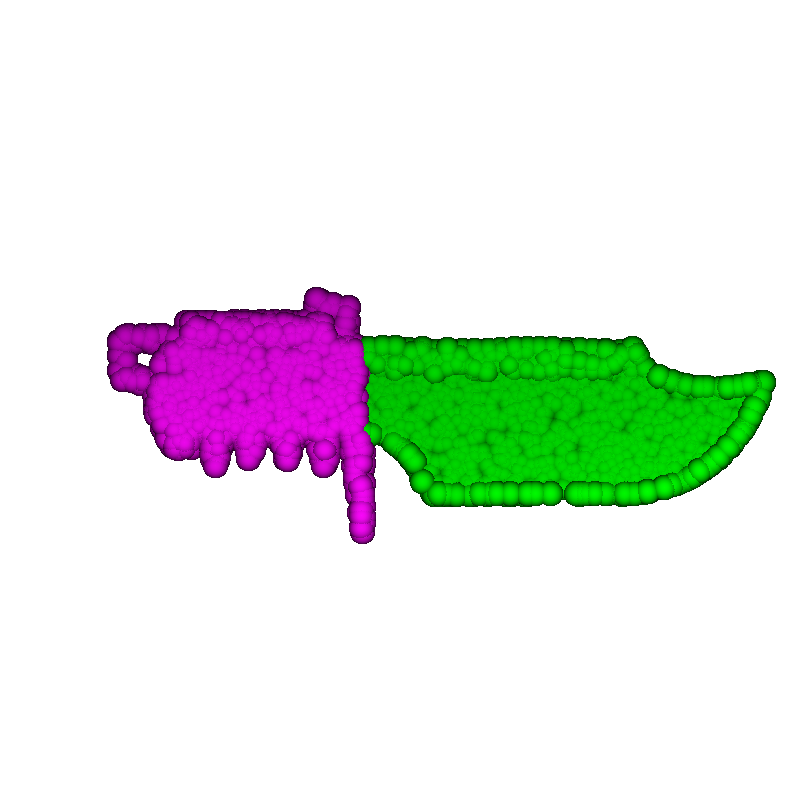}
                \caption{Knife}
                \label{fig:pred_Knife}
        \end{subfigure}
        \begin{subfigure}{0.24\linewidth}
                \includegraphics[width=\linewidth]{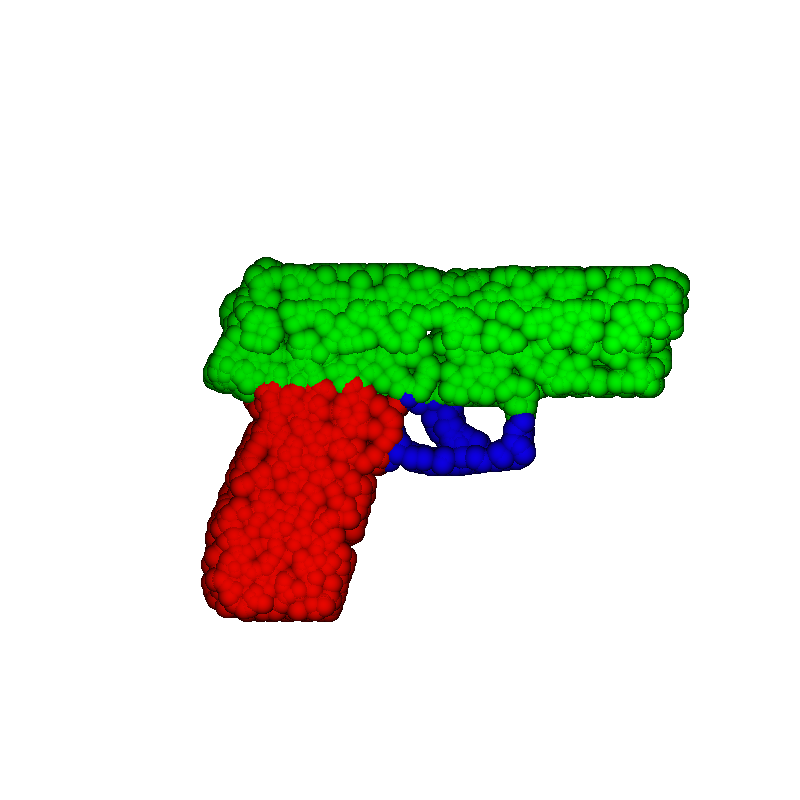}
                \caption{Pistol}
                \label{fig:pred_Pistol}
        \end{subfigure}
        \begin{subfigure}{0.24\linewidth}
                \includegraphics[width=\linewidth]{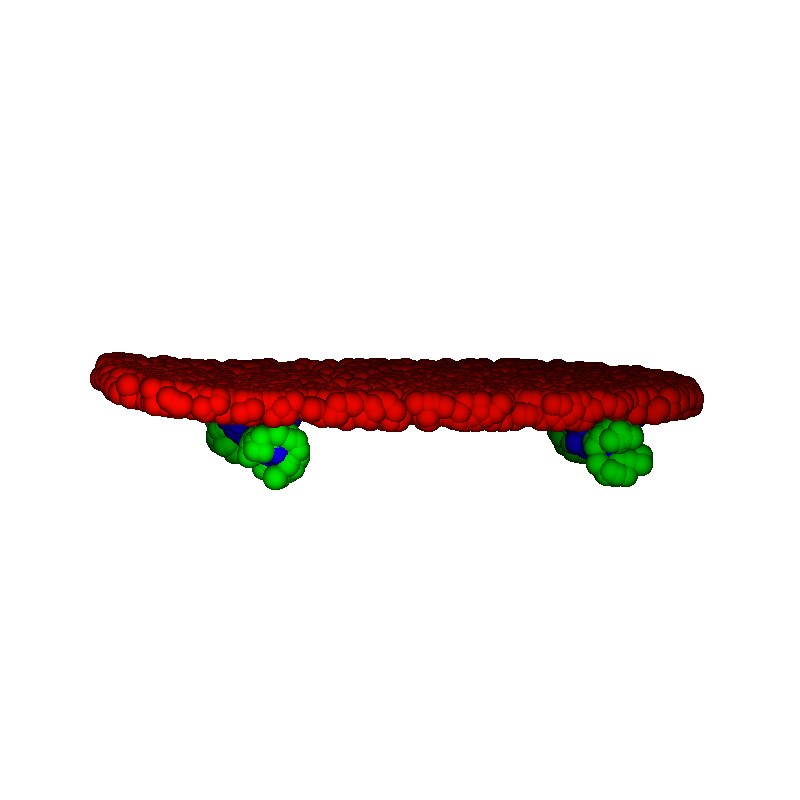}
                \caption{Skateboard}
                \label{fig:pred_Skateboard}
        \end{subfigure}
        \begin{subfigure}{0.24\linewidth}
                \includegraphics[width=\linewidth]{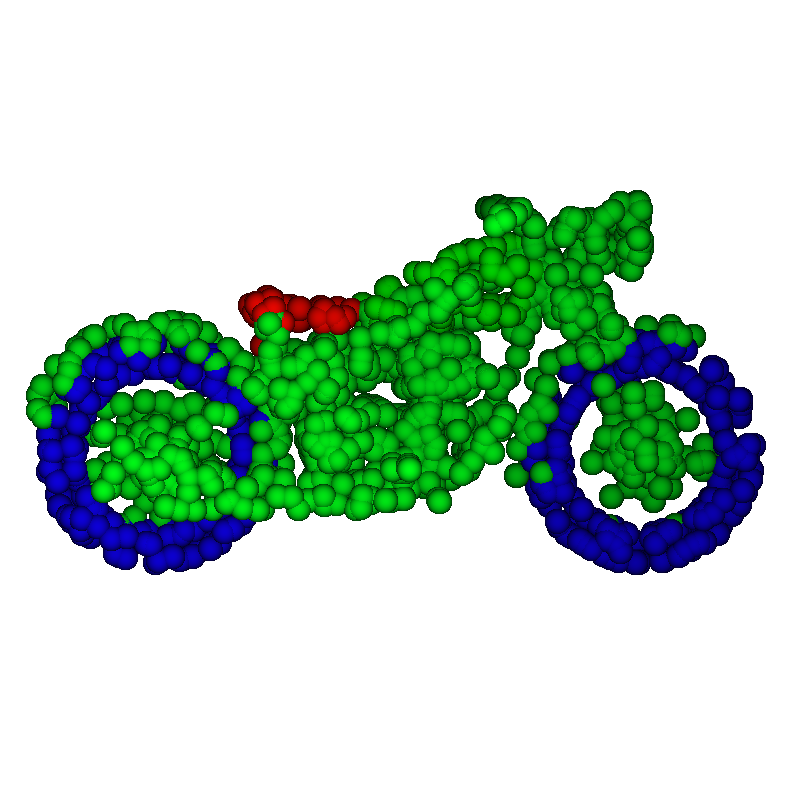}
                \caption{Motorbike}
                \label{fig:pred_Motorbike}
        \end{subfigure}
        \begin{subfigure}{0.24\linewidth}
                \includegraphics[width=\linewidth]{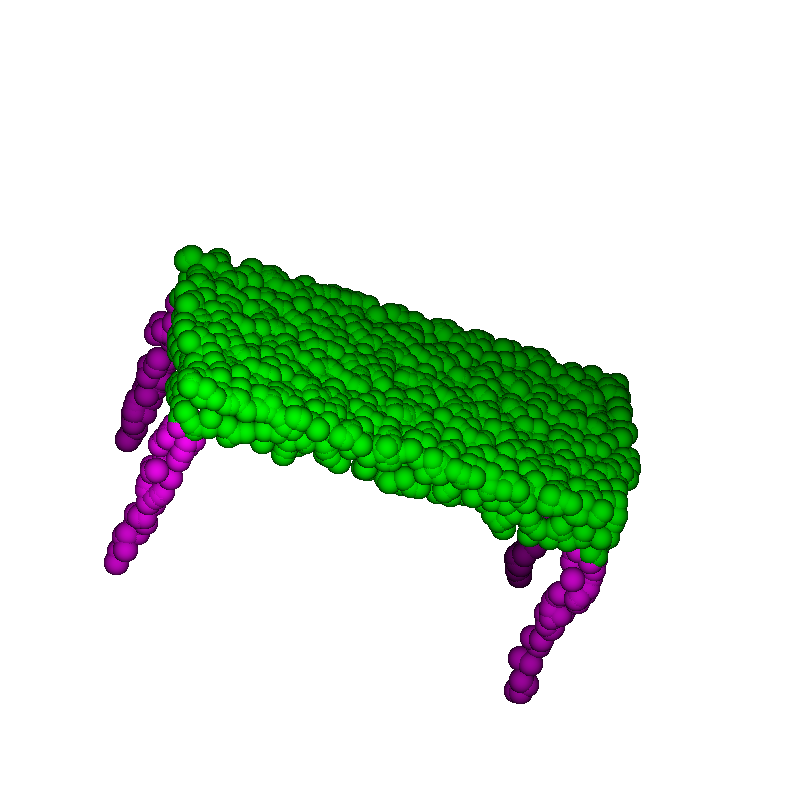}
                \caption{Table}
                \label{fig:pred_Table}
        \end{subfigure}
        \begin{subfigure}{0.24\linewidth}
                \includegraphics[width=\linewidth]{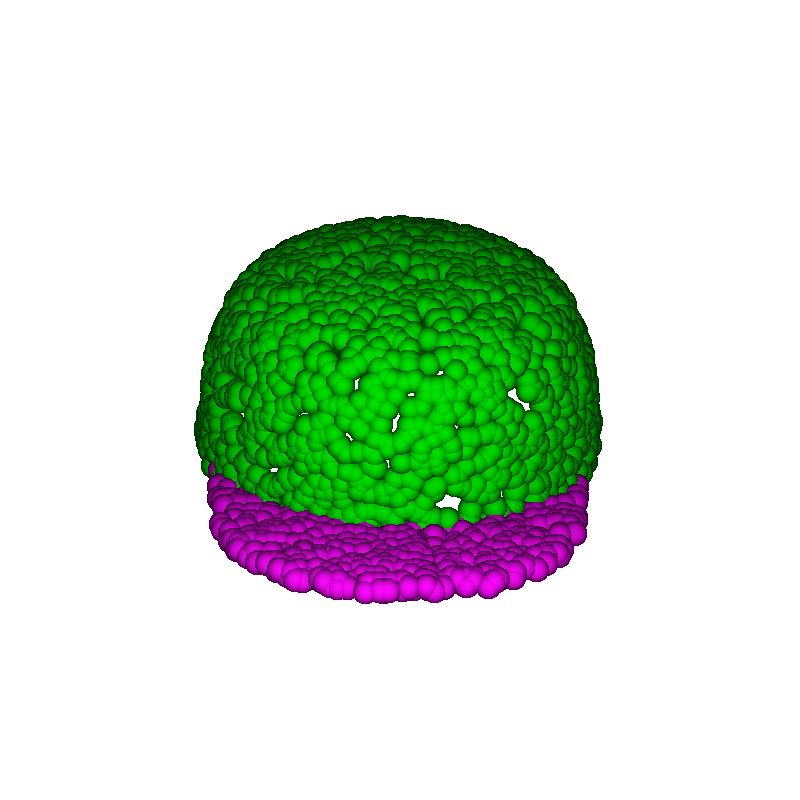}
                \caption{Cap}
                \label{fig:pred_Cap}
        \end{subfigure}
        \begin{subfigure}{0.24\linewidth}
                \includegraphics[width=\linewidth]{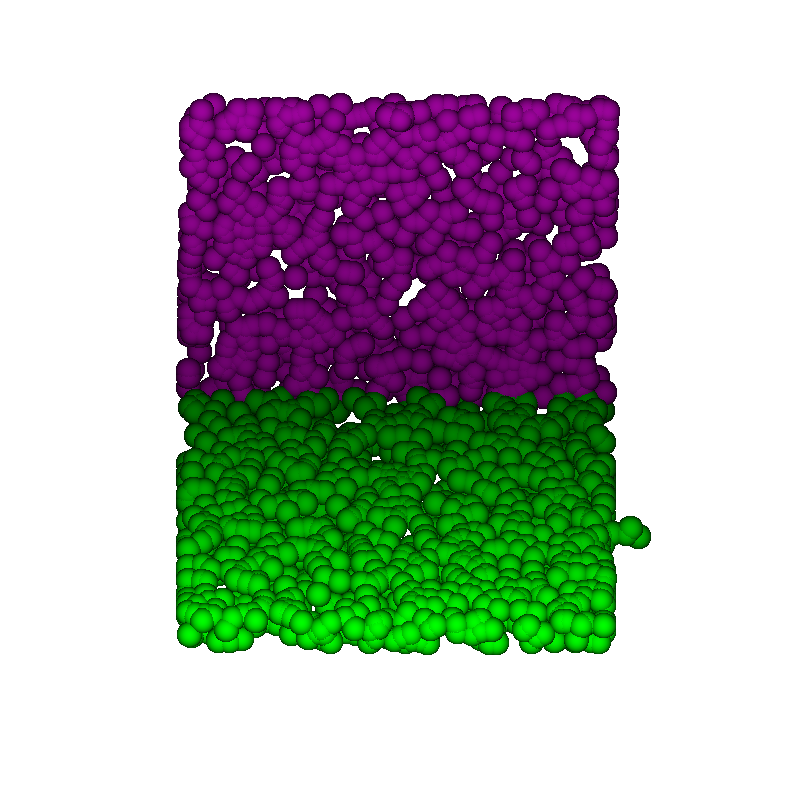}
                \caption{Laptop}
                \label{fig:pred_Laptop}
        \end{subfigure}
        \begin{subfigure}{0.24\linewidth}
                \includegraphics[width=\linewidth]{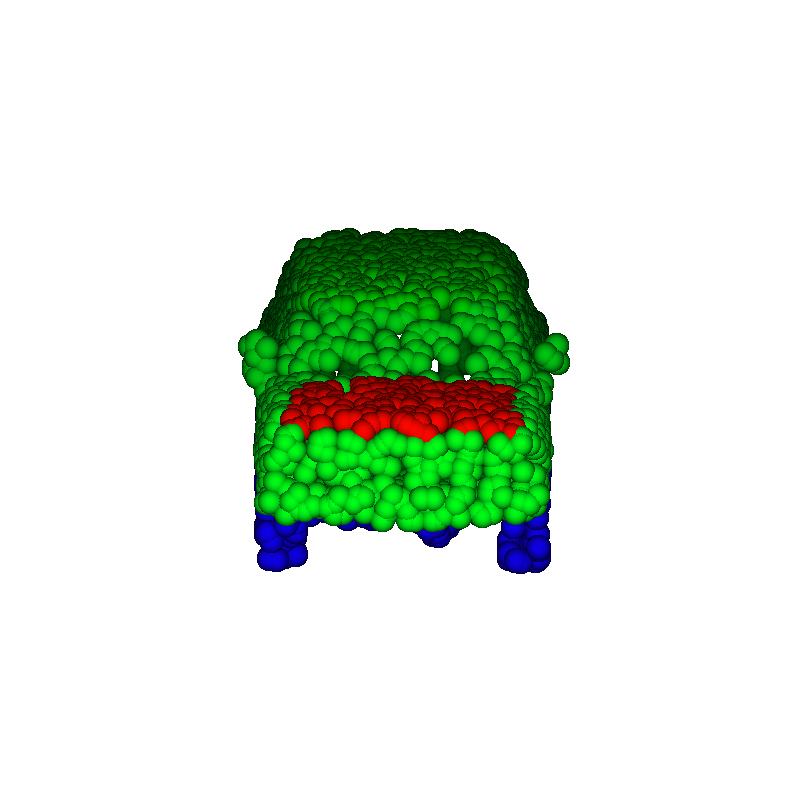}
                \caption{Car}
                \label{fig:pred_Car}
        \end{subfigure}

        \caption{Object part segmentation predictions of ViPFormer}
        \label{fig:partseg_visual}
\end{figure}

\textbf{Object Part Segmentation} 
We conduct experiments on ShapeNetPart~\cite{shapenetpart} to visualize the predictions of ViPFormer to different object parts.
This dataset has 16 object categories and we randomly select one object from each category. 
ViPFormer predicts part labels for all points in the selected object. Then different part labels are mapped to
different colors. As Fig.~\ref{fig:partseg_visual} presents, ViPFormer successfully handles different objects and segments their parts
in most cases. 

\begin{figure}[htb]
        \centering
        \begin{subfigure}{0.49\linewidth}
                \includegraphics[width=\linewidth]{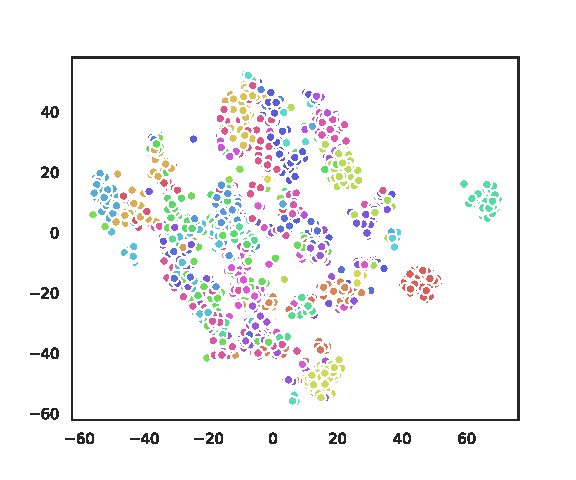}
                \caption{PT on MN, 40 categories}
                \label{fig:pt_mn_feats}
        \end{subfigure}
        \begin{subfigure}{0.49\linewidth}
                \includegraphics[width=\linewidth]{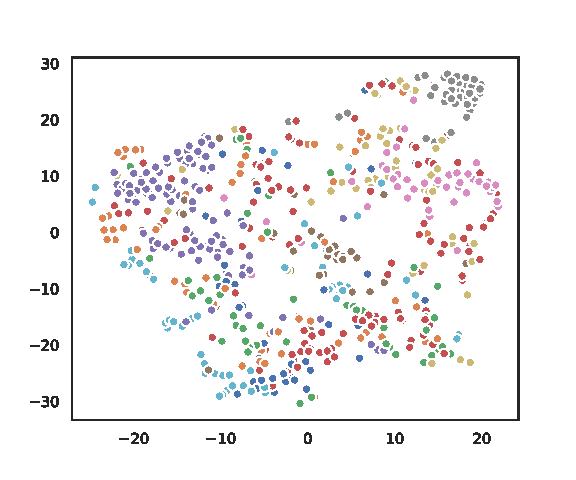}
                \caption{PT on SO, 15 categories}
                \label{fig:pt_so_feats}
        \end{subfigure}
        \begin{subfigure}{0.49\linewidth}
                \includegraphics[width=\linewidth]{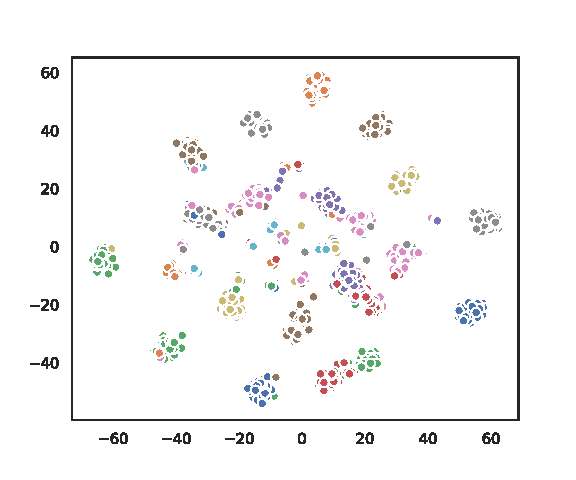}
                \caption{FT on MN, 40 categories}
                \label{fig:ft_mn_feats}
        \end{subfigure}
        \begin{subfigure}{0.49\linewidth}
                \includegraphics[width=\linewidth]{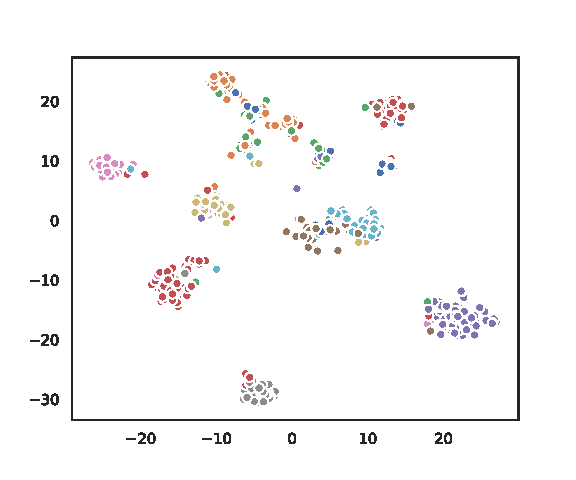}
                \caption{FT on SO, 15 categories}
                \label{fig:ft_so_feats}
        \end{subfigure}
        \caption{t-SNE~\cite{tsne} Visualization of pretrained (PT) and finetuned (FT) features on ModelNet40 (MN) and ScanObjectNN (SO).}
        \label{fig:pt_ft_feats_visual}
\end{figure}

\textbf{Feature Distribution} 
The distributions of pretrained and finetuned features are visualized by t-SNE~\cite{tsne}, exhibited in Fig.~\ref{fig:pt_ft_feats_visual}. 
The experiments are conducted on ModelNet40 and ScanObjectNN.
The pretrained features roughly scatter into different locations and provide good initialization for downstream tasks. 
After finetuning on the target datasets, the features are clearly separated by different clusters. 

\section{Conclusion}
\label{sec:conclusion}
In this paper, 
We propose an efficient Vision-and-Pointcloud Transformer to unify image and point cloud processing in a single architecture. 
ViPFormer is pretrained by optimizing intra-modal and cross-modal contrastive objectives. 
When transferred to downstream tasks and compared with existing unsupervised methods, ViPFormer shows advantages in 
model complexity, runtime latency and performances. 
And the contribution of each component is validated by extensive ablation studies. 
In the future, 
we should pay more attention to the image branch and explore its performances on downstream tasks 
since the current version focuses on point cloud understanding.   

\addtolength{\textheight}{-2cm}   





\bibliographystyle{IEEEtran}
\bibliography{IEEEabrv}

\end{document}